\documentclass[runningheads]{llncs}
\usepackage{graphicx}
\usepackage{amsmath,amssymb} 
\usepackage{color}
\usepackage[width=122mm,left=12mm,paperwidth=146mm,height=193mm,top=12mm,paperheight=217mm]{geometry}
\begin{document}
\pagestyle{headings}

\mainmatter

\title{Visual Question Generation for\\Class Acquisition of Unknown Objects} 

\titlerunning{VQG for Class Acquisition of Unknown Objects}

\authorrunning{K. Uehara et al.}

\author{Kohei Uehara${}^\text{1}$, Antonio Tejero-De-Pablos${}^\text{1}$, Yoshitaka Ushiku${}^\text{1}$, \\Tatsuya Harada${}^\text{1,2}$}


\institute{${}^\text{1}$The University of Tokyo, ${}^\text{2}$RIKEN\\
	\email{ \{uehara, antonio-t, ushiku, harada\}@mi.t.u-tokyo.ac.jp}
}

\maketitle
\begin{abstract}
Traditional image recognition methods only consider objects belonging to already learned classes. However, since training a recognition model with every object class in the world is unfeasible, a way of getting information on unknown objects (i.e., objects whose class has not been learned) is necessary. A way for an image recognition system to learn new classes could be asking a human about objects that are unknown. In this paper, we propose a method for generating questions about unknown objects in an image, as means to get information about classes that have not been learned. Our method consists of a module for proposing objects, a module for identifying unknown objects, and a module for generating questions about unknown objects.
The experimental results via human evaluation show that our method can successfully get information about unknown objects in an image dataset. Our code and dataset are available at https://github.com/mil-tokyo/vqg-unknown
\end{abstract}

\section{Introduction}

In recent years, 
in large-scale image classification tasks, image classifiers with deep convolutional neural networks (CNN) have achieved accuracies equivalent to humans~\cite{ilsvrc,resnet}. 
The recognition capabilities of these methods are limited by the object classes included in the training data. However, for an image recognition system running in the real world, for example a robot, considering all existing object classes in the world during training is unfeasible.
If such a robot was able to ask for information about {\it objects it cannot recognize}, the robot would not have to learn all classes in advance.
In this paper, we define an {\it unknown object} as an object belonging to a class not included in the training data.
In order to acquire knowledge about the unknown object class, the most reliable way is to obtain information directly from humans. For example, the robot can present an image to a human and ask them to annotate the class of an object, as in active learning~\cite{least_confident}. When the class is unknown, selecting the appropriate object and generating a suitable question about it is a challenging problem, and has not been tackled yet.
\begin{figure}[tb]
   \centering
   \includegraphics[width=0.7\hsize]{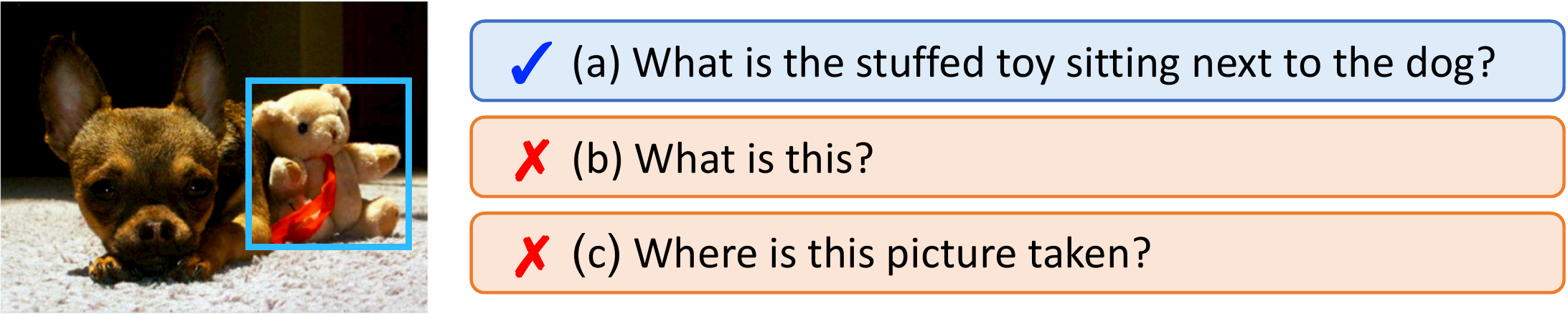}
   \caption{Examples of suitable/unsuitable questions for unknown objects. A suitable question should specify the target object ({\it stuffed toy}), so the answer is the class of the unknown object ({\it teddy bear}). Therefore, questions such as (a) are suitable.
On the other hand, simple questions such as (b) and questions about location such as (c) are unsuitable.}
   \label{fig:example}
\end{figure}
The goal of this research is to generate questions that request information about a specific unknown object in an image.
As shown in Fig.~\ref{fig:example}, compared to a simple question such as {\it ``What is this?''}, a specific question such as {\it ``What is the stuffed toy sitting next to the dog?''} targets better the class of the unknown object.
There exist several approaches~\cite{vqg1,vqg_creative} for general visual question generation (VQG) by using recurrent neural networks (RNN). Also, VQG with a specific target for a question has also been studied~\cite{disambiguate} by providing a {\it target word} (i.e., a word indicating what object the question is targeting) to an RNN as a condition. However, in these works, only the known classes are given as the target word. To the best of our knowledge, VQG targeting unknown objects has not been studied yet.
Also, in order to realize a VQG method for unknown objects, first we need to detect and classify the unknown object. However, we cannot rely on object classification methods~\cite{resnet} nor object region proposal methods~\cite{fast_rcnn} if they only consider known/labeled classes (i.e., supervised learning).

In this paper, to find unknown objects in an image, we propose object regions by selective search~\cite{selective_recognition}, which is not based on supervised learning, and then classify whether the proposed objects are unknown or not. Since our method has to classify all the objects, in order to reduce the execution time we propose an efficient unknown object classification based on uncertainty prediction.
In addition, we approach VQG for unknown objects by generating a question containing the {\it hypernym} of the unknown object.
The hypernym of a given word is another word that is higher in the semantic hierarchy, such as {\it ``animal''} for {\it ``dog.''}

{\bf Contributions}: (1) We propose the novel task of automatically generating questions to get information about unknown objects in images. (2) We propose a method to generate questions using the semantic hierarchy of the target word. (3) We construct the whole pipeline by combining modules of object region proposal, unknown object classification, and visual question generation, and show that it can successfully acquire information about unknown objects from humans.

The paper is organized as follows.
First, we explain previous studies related to this research in Sec.~\ref{related}. Next, we introduce our proposed system in Sec. ~\ref{pipeline}. Then, we show experiments on our module for unknown object classification in Sec.~\ref{unknown}, and our visual question generation module in Sec.~\ref{question}. In Sec.~\ref{all_pipeline}, we evaluate our entire pipeline to get the class of unknown objects. Finally, in Sec.~\ref{conclusion}, we discuss the conclusions and future work.

\section{Related Works}\label{related}
First, we explain active learning, an information acquisition method that also considers human help for learning. Next, we introduce the research related to each of our modules, namely, object detection, unknown object classification, and visual question generation.

{\bf Active Learning.}
The aim of active learning is achieving efficient learning by automatically selecting data that seems to contribute the most to improve the performance of the classifier and requesting a human annotator to label them.
Uncertainty Sampling~\cite{least_confident} has been proposed to select the instances whose class is the least certain.
There are three methods for Uncertainty Sampling:
(1){\it Least Confident}~\cite{least_confident}: 
Select the instance whose classification probability is the smallest and whose class has the greatest overall classification probability.
(2){\it Margin Sampling}~\cite{margin_sampling}: 
Select the instance whose difference between the most and the second most classification probabilities is the smallest.
(3){\it Entropy Sampling}~\cite{entropy1,entropy2}: 
Select the instance whose distribution of classification probabilities has the largest entropy.
The main difference between active learning and this research is that active learning targets only instances whose class is included in the training set, whereas we target instances whose class is not in the training set. 
Also, active learning only presents data to the annotator; it does not generate questions.

{\bf Object Region Proposal.}
Object region proposal methods detect the region surrounding objects in an image.
Recent methods perform object detection that performs both object region proposal and object classification at the same time via supervised learning using CNN~\cite{fast_rcnn,yolo}.
These methods achieve accurate object detection with a huge amount of labeled data for training. However, they do not consider unknown objects.
In contrast, there is some research on {\it objectness} that simply estimates the existence objects in a specific region of the image, without classifying the object. Alexe et al.~\cite{objectness} perform objectness estimation by  using saliency, contrast, edge, and superpixel information. Cheng et al.~\cite{bing} learn objectness from image gradients.
Also, a method called selective search~\cite{selective_recognition,selective_segmentation} allows object region proposal by using image segmentation, and integrating similar regions with each other. Since it does not require object labels, it can propose regions without learning the object class.

{\bf Unknown Object Classification.}\label{classification}
Unknown object classification performs binary classification of objects in an image as {\it known} or {\it unknown}.
Traditionally, object classification methods estimate the actual class of an object in an input image.
Recent research in object classification are CNN-based methods~\cite{resnet,vgg}.
These methods assume a {\it closed set}, that is, they only consider the classes included in training and not unknown classes.
On the other hand, there is research on the task called {\it open set recognition}~\cite{openset} for object classification that includes unknown objects.
Open set recognition is a task aimed at classifying to the correct class if the input belongs to the trained class, and if the input is unknown, classifying it as unknown.
For open set recognition, methods using SVM~\cite{openset} and methods extending the nearest neighbor method~\cite{openworld} have been proposed.
Also, Bendale et al.~\cite{openmax} proposed open set recognition using CNN. They classify an object as unknown if its feature distribution extracted from the CNN hidden layers is distant from known classes.

{\bf Visual Question Generation.}
Visual Question Generation (VQG) was recently proposed as an extension of image captioning. Whereas image captioning methods~\cite{showandtell} generate descriptive sentences about the content of an image, VQG methods generate questions (e.g., {\it What color is the car?}).
The common approach in VQG is encoding image features via CNN and generating a sentence by decoding those features using an RNN. Methods that use a gated recurrent unit (GRU)~\cite{vqg1} and a long short-term memory (LSTM)~\cite{vqg_creative} have been proposed.
Traditional VQG methods generate questions from the whole image, without focusing in any particular image region.
Only recently, methods that generate questions targeting a particular image region have been proposed.
Zhang et al.~\cite{groundedvqg} detect different regions to generate a variety of questions from the same image.
In contrast, Li et al.~\cite{disambiguate} generate questions focusing on a specific region with the goal of distinguishing between two images. For this, they input a target word (e.g., blue) related to the region as a condition to the LSTM. In~\cite{disambiguate}, target words are known classes learned in advance. To the best of our knowledge, VQG targeting an unknown object has not been approached yet.

\section{Proposed System}
\label{pipeline}

\begin{figure}[tb]
   \centering
   \includegraphics[width=\hsize]{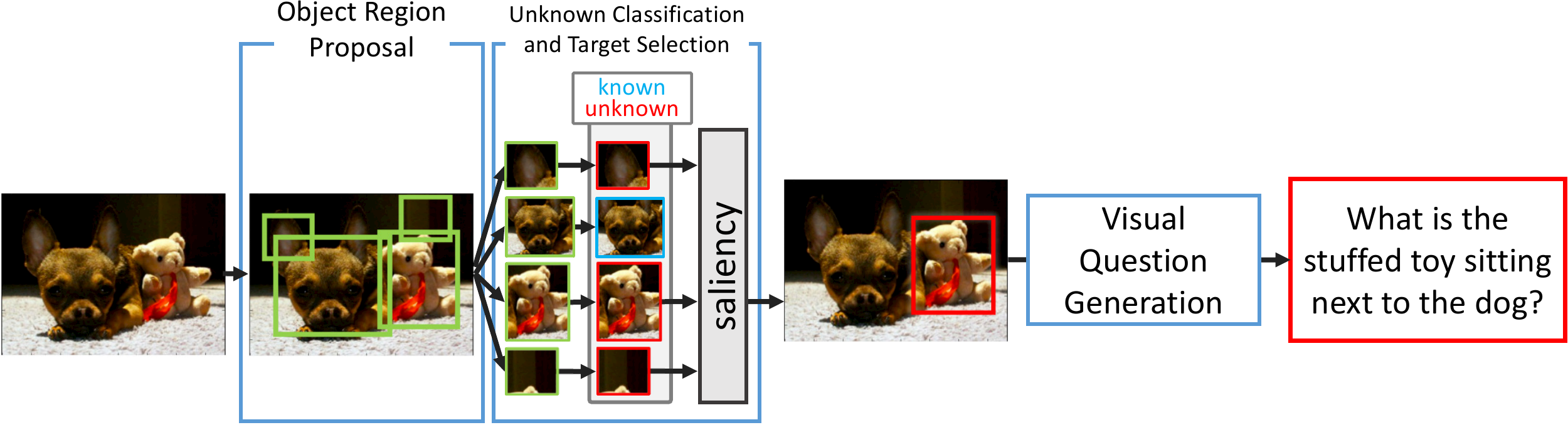}
   \caption{Overview of the proposed method.
First, regions from objects in the image (including unknown objects) are detected.
Then, unknown objects are classified and the target region is selected.
Finally, the target region along with the whole image is coded into a feature vector, and a question for the unknown object is generated}
   \label{fig:all_pipeline}
\end{figure}

Fig.~\ref{fig:all_pipeline} shows the overview of the proposed method.
First, objects in the input image are detected by the object region proposal module.
Next, the unknown object classification and target selection module identifies whether each object is unknown or not, and selects an object region to be the target of the question. We refer to this region as the {\it target region}.
Finally, the visual question generation module generates a question using features extracted from the whole image and the target region.

\subsection{Object Region Proposal}
Our object region proposal module detects all objects in the input image via selective search.
The proposed method needs to detect unknown objects (i.e., objects never learned before), so supervised learning is not an option since it requires labels for all objects.
As mentioned in Sec.~\ref{related}, selective search provides candidate regions for objects without supervised learning. Thus, unknown objects can also be detected, and the number of object regions can be reduced compared with an exhaustive search.
Therefore, this seems to be suitable as a method for object region proposal.

\subsection{Unknown Object Classification and Target Selection}

This module selects the {\it target object}, that is, the object to acquire information about. For that, we classify objects into {\it known} or {\it unknown}, and then select the most salient unknown object. This prevents generating questions about unimportant regions that may have been proposed by mistake by the object region proposal module.
We define unknown object classification as follows: for an input object image, if its class is included in the training set, classify it to the correct class, and if not, classify it as unknown.
Specifically, we perform unknown object classification on the classification results of a CNN as follows.
The output of the softmax function of the CNN can be regarded as the confidence with which the input is classified into a certain class. 
We consider that images of unknown objects result in a low confidence value for all classes.
That is, the more uniform the confidence distribution, the lower the confidence for all classes and the more possibilities the object is unknown.
Therefore, we perform unknown object classification by estimating the dispersion of the probability distribution using an entropy measure, with reference to the method of Uncertainty Sampling in active learning~\cite{entropy1,entropy2}.
The entropy measure $E$ is defined as:
\begin{equation}
E = - \sum_{j=1}^K p_j \log_2 p_j
\end{equation}
where $p_j$ is the output of the softmax function when a given input $x$ is classified into class $C_j$ $( j = 1, 2, ..., K )$.
$E$ takes the maximum value $\log_2K$ when all $p_j$ are all equal, that is, when $p_j = 1/K$. On the other hand, the larger the dispersion of the probability distribution is, the smaller the entropy becomes.

Also, it is necessary to select which object to generate a question about among the objects classified as unknown.
For example, in some cases, the region proposed by selective search contains only the background.
Background regions are likely to be classified as unknown, but they do not contain an object to ask about.
In order to solve this problem, we calculate the saliency of each proposed region in the image as a criterion for selecting the target region.
That is, we ask questions about objects that are unknown and particularly salient in the image.
Thus, to select salient objects in the image, we propose using a saliency map.
The saliency map is a plot obtained by estimating the saliency for each pixel in the image.
We calculate the saliency map using the method of Zhu et al.~\cite{saliency}.
This method estimates low saliency for background pixels and high saliency for foreground pixels. Therefore, it is considered to be suitable for this research.
First, we preprocessed the image by applying mask based on saliency map and applied non-maximum suppression to reduce the large number of object regions.
Then, the saliency of each proposed object region is expressed by:

\begin{equation}
I_{region} = \sum_{I(p)\; \ge \; \theta}I(p) \times \frac{S_{salient}}{S_{region}}
\end{equation}
where, $I(p)$ is the saliency value of each pixel, $\theta$ is the threshold value, $S_{salient}$ is the area in the region where saliency exceeds $\theta$, and $S_{region}$ is the total area of the region. The threshold $\theta$ was determined using Otsu method~\cite{otsu}.
The region with the highest saliency is selected as the target region.

\subsection{Visual Question Generation}\label{proposal:question}

\begin{figure}[tb]
   \centering
   \includegraphics[width=0.7\hsize]{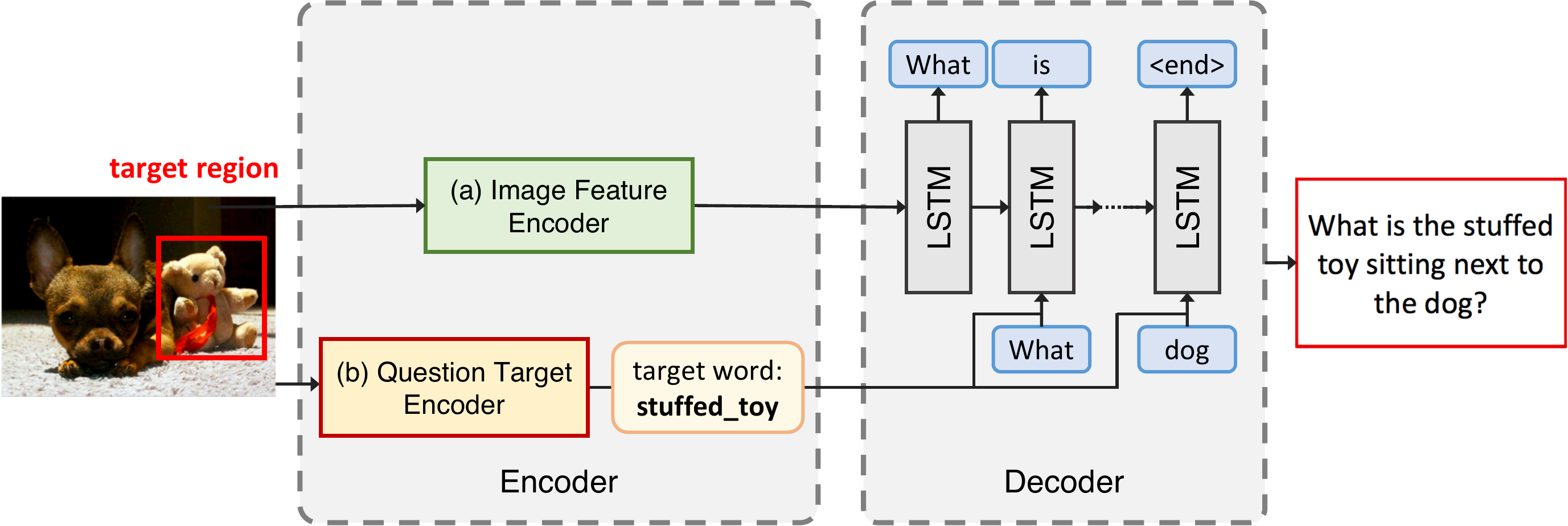}
   \caption{Overview of the VQG module. First, we obtain the common hypernym of the prediction class of the classifier as the target word. The target word and the image features are input as conditions to the LSTM, and the question is generated}
   \label{fig:question_module}
\end{figure}

Figure~\ref{fig:question_module} depicts the visual question generation module. We generate a question following the encoder-decoder methodology of Mostafazadeh et al.~\cite{vqg1} and Li et al.~\cite{disambiguate}. The encoder extracts visual features of both the entire image and the object region (submodule (a)), and (submodule (b)) the target word into a word embedding vector representation. The decoder takes the encoded features and generates a question via LSTM.

{\bf Encoding of Image Features.}
This submodule uses a pretrained CNN model to extract the features $f_I$ of the entire image and the features  $f_R$ of the target region.
In our method, we use the output (a 1,000-dimensional vector) of the {\it fc} layer of ResNet152~\cite{resnet}.
Then, in order to express the spatial information of the target region, we follow the method by Li et al.~\cite{disambiguate} to define a five-dimensional vector $l_R$ as:
\begin{equation}
l_R = \left[\frac{x_{tl}}{W}, \:\frac{y_{tl}}{H}, \:\frac{x_{br}}{W}, \:\frac{y_{br}}{H}, \:\frac{S_R}{S_I} \right]
\end{equation}
where $(x_{tl}, \:y_{tl}) $, $(x_{br}, \:y_{br})$ is the upper left and the lower right coordinate of the target region, $S_R$ and $S_I$ represent the area of the target region and the entire image respectively, and $W$ and $H$ denote the width and the height of the image respectively.
We concatenate $f_I,\:f_R,\:l_R$, and let the 2,005 dimensional vector $f = \left[f_R, f_I, l_R \right]$ be the image feature encoding.

{\bf Question Target.}\label{proposed:question}
This submodule selects a target word to represent the object in the target region and embeds it into a vector representation.
Since the target object is unknown, that is, is not in the trained classes, it is not possible to use the class label as the target word as in Li et al.~\cite{disambiguate}.
Therefore, we need to devise how to specify the target word.
For example, if we do not know the class dog, asking a question referring to an {\it animal} is natural (e.g., {\it ``What is this animal?''}).
In this case, the word {\it animal} for {\it dog} is considered to be a hypernym. Such hypernym can be used as the target word.
We use WordNet~\cite{wordnet} to get the hierarchical relationship of words.
Each word in WordNet is hierarchically arranged based on semantic relationships, and thus, it is possible to get the hypernym of a word by going up in the hierarchy.

\begin{figure}[tb]
   \centering
   \includegraphics[width=0.8\hsize]{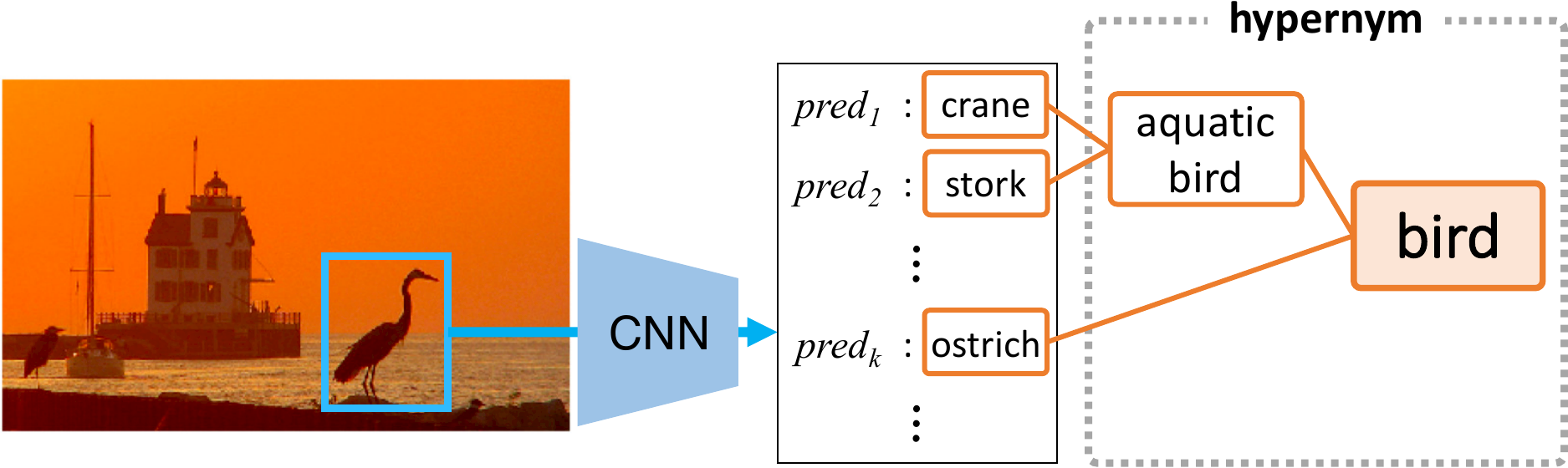}
   \caption{Overview of the question target module for target word selection using WordNet~\cite{wordnet}. WordNet is used to obtain a hypernym common to the predicted class labels with the highest confidence. The hypernym is then input to the visual question generation module}
   \label{fig:q_target}
\end{figure}

As shown in Fig.~\ref{fig:q_target} , we use the $ k $ predicted classes ($pred_1, pred_2, \dots pred_k$) with the highest confidence of the classification result, and we select the word with the lowest level among the common hypernyms of the $k$ class labels.
If the value of $k$ is too large, the common hypernym becomes a very abstract word such as {\it whole} or {\it entity}, and it is not possible to designate the target appropriately.
Therefore, the value of $k$ should be chosen carefully.

Then, we use the Poincar\'e Embeddings~\cite{poincare} to embed words into feature vectors using a neural network similar to Word2Vec~\cite{word2vec1,word2vec2}. However, unlike Word2Vec, Poincar\'e Embeddings are suitable for expressing a structure in which words are hierarchically represented, such as WordNet, as a vector.
Let the target word embedded by Poincar\'e Embeddings be the vector $\sigma(v)$. Then the input to the decoder LSTM is the visual feature vector $f$, and the conditional input is the word embedded vector $\sigma(v)$.
The decoder LSTM is trained by minimizing the negative log likelihood:
\begin{equation}
L = \sum -\log p(Q \;|\;f, \;\sigma(v)\;;\; \theta) \label{eq:target_loss}
\end{equation}
where $\theta$ is the parameters of the LSTM, and $Q$ denotes the generated question.

\section{Evaluation of the Unknown Object Classification}\label{unknown}

Before evaluating the entire pipeline we performed experiments on independent modules to study their performance. First, we evaluated how accurately the unknown object classification module can classify whether the input image is an unknown object image or not.
\subsection{Experimental Settings}

We used CaffeNet~\cite{alexnet}, VGGNet~\cite{vgg}, and ResNet152~\cite{resnet}, which are well-known CNN models, to study the variation in unknown classification accuracy when employing different classifiers.
We pretrained our classifier with the 1,000 class dataset used in the object classification task of ILSVRC2012.
We used 50,000 images of the same dataset for validation.
Then, we used the dataset in the object classification task of ILSVRC2010 to create the unknown dataset. We excluded all images whose class is {\it known} (i.e., included in the ILSVRC2012 dataset), as well as the images whose class is a hypernym of any {\it known} classes. The reason for removing hypernyms is to avoid including general classes (e.g., ``dog'') in the unknown dataset when the specific class is already known (e.g., ``chihuahua'').
Thus, we selected 50,850 images of 339 classes from ILSVRC2010 dataset, which are not included in ILSVRC2012 and its hypernyms.

\subsection{Methods}
We compare unknown object classification using entropy $E$, which is the proposed method, with the following two methods.

{\bf Least Confident}~\cite{least_confident}
We used the method of Uncertainty Sampling described in Sec.~\ref{related}. We set a threshold for the softmax probability of the class label with the highest probability, that is, the most probable label. Then, if the probability is lower than the threshold, the label is considered as unknown.

{\bf Bendale et al.}~\cite{openmax}
We used the method of Bendale et al. mentioned in Sec.~\ref{classification}.
We performed this experiment based on the code published by the authors, and change the classification models for CaffeNet, VGGNet, and ResNet.

\subsection{Evaluation Metrics}
We calculated the F measure as:
\begin{equation}
F = \frac{2TP}{2TP+FP+FN}
\end{equation}
where $TP$ is defined as the number of known data classified into the correct class, $FN$ as the number of unknown data misclassified into known data, and $FP$ as the number of misclassified known data~\cite{openmax}.
We performed the evaluation using a five-fold cross-validation.

Also, we measured the execution time of each method.
First, we measured the time required by the classifier to calculate the softmax probability distribution of one image. Next, we experimented the calculation time per image for each method, taking the distribution of softmax probability for 100 images as input.
We repeated this operation five times and calculated the average execution time.

\subsection{Experimental Results}
Table~\ref{table:f_result} shows the resulting F measure per classifier and method.
\begin{table}[t]
\centering
\caption{Comparison of the proposed unknown object classification method in terms of F measure results $\pm$ standard error. We performed experiments on CaffeNet, VGGNet, and ResNet. In all three cases, the proposed method outperformed the other methods}
\label{table:f_result}
\begin{tabular}{c|c|c|c}
\hline
 & \multicolumn{3}{c}{F measure} \\ \hline
 & CaffeNet & VGGNet & ResNet \\ \hline \hline
Ours & ${\bf 0.526\pm1.1\cdot10^{-3}}$ & ${\bf 0.602\pm0.2\cdot10^{-3}}$ & ${\bf 0.654\pm0.9\cdot10^{-3}}$ \\ \hline
Least Confident & $0.522\pm1.1\cdot10^{-3}$ & $0.590\pm1.5\cdot10^{-3}$ & $0.635\pm1.2\cdot10^{-3}$ \\ \hline
Bendale et al.~\cite{openmax} & $0.524\pm0.9\cdot10^{-3}$ & $0.553\pm0.6\cdot10^{-3}$ & $0.624\pm1.7\cdot10^{-3}$ \\ \hline
\end{tabular}
\end{table}
\begin{table}[t]
\caption{Comparison of the proposed unknown object classification method in terms of execution time, with CaffeNet as a classifier. We performed classification for 100 images and showed the average time per image $\pm$ standard error}
\label{table:time}
\begin{center}
\begin{tabular}{c|c}
\hline
&time (sec/image)\\
\hline\hline
Ours&$0.0400\pm0.0017$\\
\hline
Least Confident&${\bf 0.0365\pm0.0019}$\\
\hline
Bendale et al.~\cite{openmax}&$15.6\pm0.7$\\
\hline
\end{tabular}
\end{center}
\end{table}
For all three methods, the F measure increased as the classifier was changed from CaffeNet to VGGNet, and to ResNet.
The higher the accuracy of the classifier, 
when inputting known classes, 
the distribution of the classification probabilities varies more largely, and thus the entropy becomes smaller.
In the case of inputting unknown classes,  the more accurate the classifier is, the distribution of the classification probabilities is more uniform and thus the entropy becomes larger.

Table~\ref{table:time} shows a comparison of the execution time for each method when CaffeNet is used as a classifier. Our method and the Least Confident method take much less time than the method of Bendale et al.
This is because the method of Bendale et al. has to calculate the distance to the average distribution of 1,000 known classes for each image, so the calculation cost is large, but in the method using entropy and the method using threshold of confidence, calculation is performed only with distribution of the input image, so calculation time is shortened.

\section{Evaluation of the Visual Question Generation}\label{question}
We studied the performance of the proposed visual question generation module given a target region and compared to other methods.

\subsection{Datasets}\label{unknown_dataset}
In this experiment, we used a dataset called Visual Genome~\cite{visual_genome} with about 100,000 images, and captions and questions.
There is a subset of questions that is associated with a specific region in the image.
We preprocessed the data as follows.
First, 
we removed questions not beginning with {\it ``What.''}
Furthermore, since questions about colors are not the goal of our method, we also removed questions beginning with {\it ``What color.''}
Next, for questions associated with a specific region in the image, if the word representing the object in the region was included in the answer of the question, that word was taken as the target word.
For questions not associated with an image region, we searched the object included in the answer among all objects in the image. Then, if there is only one instance of the object in the image, the word and the region where the object exists are set as the target word and the target region corresponding to the question.
Furthermore, in order to eliminate the imbalance in the type of questions in the data, we limited to 50 the maximum number of times the same question can be included.
Through this preprocessing, we gathered 202,208 questions corresponding to specific target regions in the image (one question per region) and 528 target words.

\subsection{Methods}\label{experimental}
We split the 202,208 questions into 179,997 questions for training, 10,002 questions for validation, and 12,209 questions for testing. At training time, questions were generated by inputting images, regions, and target words. For embedding target words, we used Poincar\'e Embeddings trained on the tree structure of WordNet.
We used the following methods as a baseline to compare our proposed method.

{\bf CNN + LSTM.}
As in Mostafazadeh et al.~\cite{vqg1}, we generated questions by inputting only the features of the entire image encoded by a CNN.

{\bf Retrieval}.
Following Mostafazadeh et al.~\cite{vqg1}, we also used retrieval method as baseline. First, we extracted features of the target regions in the training images using the {\it fc} layer of ResNet152. Then we retrieved the $m$ regions with the higher cosine similarity between their features and the input target region.
Then, for each question associated to the retrieved region, we calculated the similarity with the other $m-1$ questions using the BLEU score~\cite{bleu}, which measures textual similarity.
Finally, the question with the highest BLEU score, that is, the most representative question, was taken as the final output.

\subsection{Evaluation Metrics}
In our experiments, we use BLEU~\cite{bleu} and METEOR~\cite{meteor} for measuring the similarity between the automatically generated questions and the ground truth. The larger the value, the more accurate the result.

Besides the automatic evaluation, we also performed human evaluation via Amazon Mechanical Turk (AMT)\footnote{https://www.mturk.com/}.
We presented an image with the target region and the target word to the human workers. We asked workers to blindly evaluate each method and the ground truth using a score between 5 (best) and 1 (worst). We used two criteria for evaluation: (1) whether each question is expressed naturally, and (2) whether each question is related to the target region and the target word.
For the human evaluation, we used questions generated for 100 images extracted randomly from the test data.

\begin{table}[tb]
\caption{Comparison between our method and the baseline in terms of automatic evaluation metrics. The proposed method outperformed baseline methods}
\label{table:auto_result1}
\begin{center}
\begin{tabular}{c|ccccc}
\hline
&BLEU-1&BLEU-2&BLEU-3&BLEU-4&METEOR\\
\hline\hline
Ours&{\bf0.518}&{\bf0.359}&{\bf0.244}&{\bf0.175}&{\bf0.197}\\
\hline
CNN + LSTM&0.456&0.296&0.175&0.110&0.163\\
\hline
Retrieval&0.438&0.275&0.157&0.094&0.151\\
\hline
\end{tabular}
\end{center}
\end{table}
\begin{figure}[tb]
  \begin{minipage}[b]{0.47\linewidth}
    \centering
    \includegraphics[width=\linewidth]{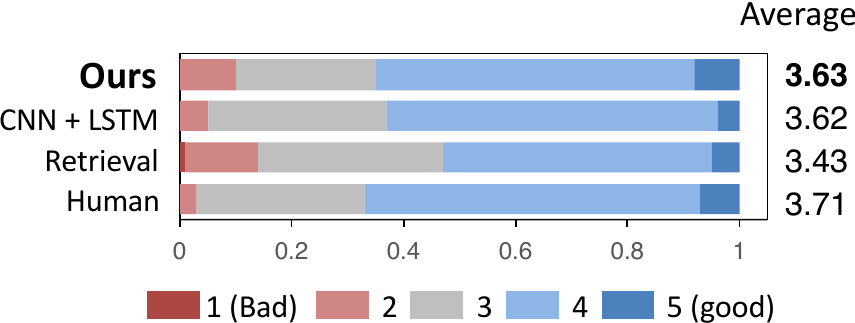}
    \caption{(1) Human evaluation results on the naturalness of questions }\label{fig:q_amt_natural}
  \end{minipage}
  \begin{minipage}{0.02\hsize}
        \hspace{2mm}
      \end{minipage}
  \begin{minipage}[b]{0.47\linewidth}
    \centering
    \includegraphics[width=\linewidth]{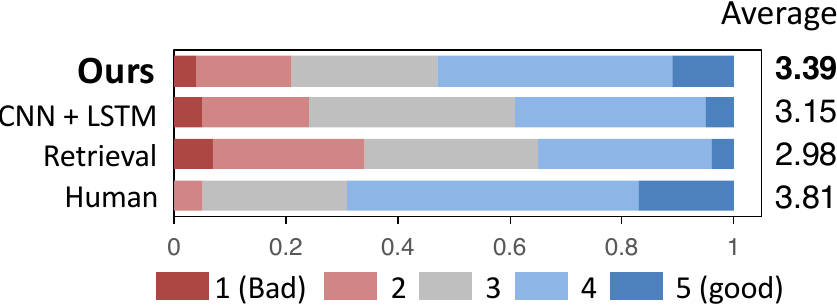}
    \caption{(2) Human evaluation results on the relevance of questions to their region}\label{fig:q_amt_ref}
  \end{minipage}
\end{figure}

\subsection{Experimental Results}
As shown in Table~\ref{table:auto_result1}, the proposed method outperformed the baselines for all metrics.
This result suggests that inputting the target object (visual features and target word condition) to the decoder LSTM allows generating more accurate questions.
Figures~\ref{fig:q_amt_natural} and \ref{fig:q_amt_ref} show the results of the human evaluation of our method compared to baselines. 
From the viewpoint of the naturalness of the question, the difference between the proposed method and CNN + LSTM was small.
We believe the reason is that both methods use LSTM as the decoder.
When evaluating the relevance of the question to the target region, the proposed method outperformed baselines. This is because, CNN + LSTM does not specify a target object to the decoder, so it may generate questions that are related to the image but not the target region.
Also, Retrieval can generate only questions existing in the training data, so the variety of questions is limited, and thus, it may not generate questions related to the target region.

\section{Evaluation of VQG for Unknown Objects}\label{all_pipeline}
Lastly, we performed experiments using the whole pipeline, in which we generate questions to acquire knowledge about the class of unknown objects in the image.

\subsection{Datasets}
In order to test our VQG method for unknown objects, we used images that include unknown objects extracted from the following two datasets.
First, from the test set of Visual Genome, we extracted 50 images with unknown objects, that is, not included in the 1,000 classes of ILSVRC2012.
Also, from the dataset of the ILSVRC2010, 50 images of 339 unknown classes as described in Sec.~\ref{unknown} were extracted.
In the images from the Visual Genome dataset, target regions contain an average of 8.7 objects, including small objects like ``eye'' and ``button''. According to our method, 68.4 \% of those objects were unknown. Note that we cannot indicate objectively the number of objects in the images from the ILSVRC2010 dataset since its ground truth does not include object regions.

\subsection{Methods}
The classifier used for unknown object classification was ResNet152, which is the method with the highest accuracy in Sec.~\ref{unknown}. We pretrained ResNet152 with the 1,000 class data used in the object classification task of ILSVRC2012. The visual question generation module was pretrained with the dataset created in Sec.~\ref{unknown_dataset}.
Furthermore, as described in Sec.~\ref{proposed:question}, when choosing a hypernym common to the top $k$ classification results, if the value of $k$ is too large, the target word becomes too abstract.
Therefore, we performed experiments with two settings, $k=2$ and $k=3$.
As baseline methods, we used the CNN + LSTM method and the Nearest Neighbors Retrieval method described in Sec.~\ref{question}.

\subsection{Evaluation Metrics}
Since there is no ground truth in this experiment, it is not possible to perform automatic evaluation by comparison with the ground truth. Therefore, we performed only human evaluation via AMT, which consists of the following two tasks.

(1) We presented to three workers images and the questions generated automatically by our method and the baselines, and asked them to answer the generated questions. When they cannot understand the meaning of the question, we instructed them to answer {\it ``Do not understand.''} Note that this task did not present a target region.

(2) Also, we evaluated the question and the answer obtained in task (1).
Specifically, we presented to three workers with the question, the answer of each worker in task (1), and the image with the target region, and asked them whether the question and the answer are related to the target region in a 5-point scale.
We evaluated only answers different from {\it ``Do not understand.''}

As the evaluation value for task (2), we used the median of the evaluation values of the three workers.

Lastly, we evaluated to what extent the generated questions are able to successfully acquire information on unknown objects. We counted only the questions whose answers (task (1)) are not included in the known classes of the classifier, and the relevance of the question and target region in the image (task (2)) is four or more.

\subsection{Experimental Results}

\begin{figure}[tb]
\centering
\includegraphics[width=\linewidth]{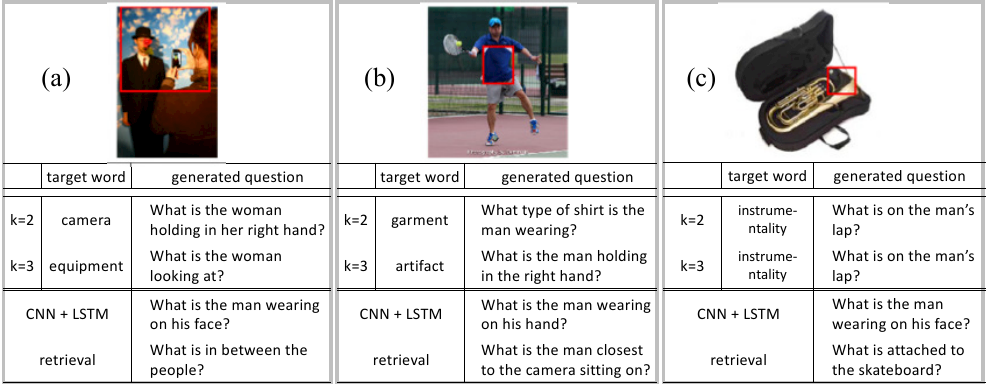}
\caption{Examples of input images (upper), the target words and generated questions by our proposed VQG method for unknown objects (middle), and the generated questions by the \textit{CNN + LSTM} and \textit{retrieval} baselines (lower).}
\label{fig:all_result}
\end{figure}

Fig.~\ref{fig:all_result} shows our qualitative results.
In Fig.~\ref{fig:all_result} (a) and (b), when $k=2$, the target word is a concrete word (i.e., {\it ``camera''} and {\it ``garment''}), and the generated question refers to an object in the target region.
In the case of $k=3$, the target word is an abstract word such as {\it ``equipment''} and {\it ``artifact''}, and the generated question is not related to the region.
Fig.~\ref{fig:all_result} (c) shows an example where the object region proposal is not performed properly, and thus, it is not possible to generate the question accurately.
The lower part of the image shows examples of questions generated by the baselines.

\begin{figure}[tb]
\begin{tabular}{lcr}
\begin{minipage}{0.50\textwidth}
   \centering
   \includegraphics[width=1.0\hsize]{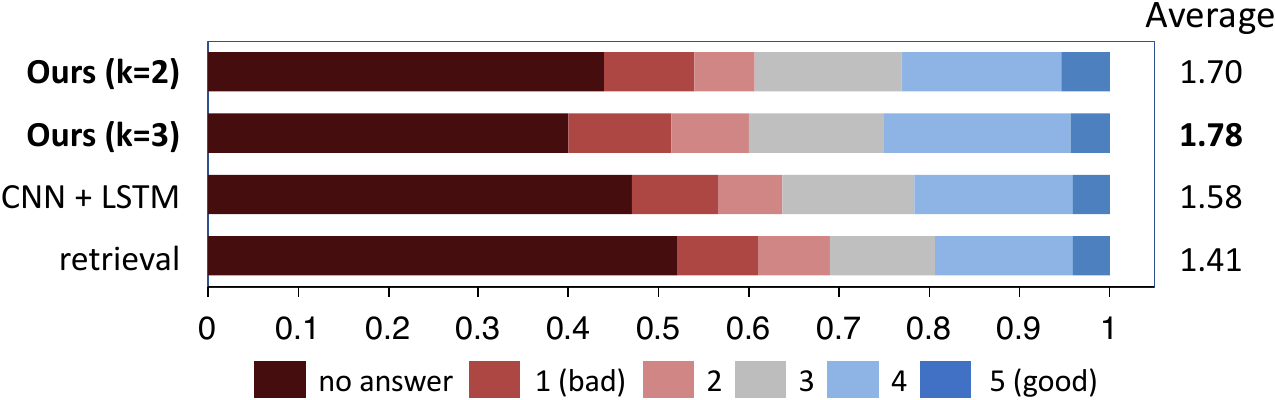}
   \caption{Comparison of our method with the baseline in terms of the human evaluation in task (2). Task (2) evaluates whether or not the generated question, the image region, and the obtained answer are related. The greater the score, the higher the relevance.}
   \label{fig:amt_all1}
 \end{minipage}&
 \begin{minipage}{0.02\textwidth}
 ~
 \end{minipage}
 \begin{minipage}{0.48\textwidth}
      \centering
      \makeatletter
        \def\@captype{table}
      \makeatother
      \caption{The number of generated questions that successfully allowed acquiring information on unknown objects (out of 300). We counted only the questions whose answers (task (1)) are not included in the known classes of the classifier, and the relevance of the question and target region in the image (task (2)) is four or more.}
      \label{table:unknown}
      \begin{tabular}{c|c}\hline
      Ours($k=2$)    & {\bf 61}      \\
      Ours($k=3$)    & 49      \\
      CNN + LSTM & 46      \\
      Retrieval     & 45      \\ \hline
      \end{tabular}
\end{minipage}
\end{tabular}
\end{figure}

Figure~\ref{fig:amt_all1} shows the results of the human evaluation. The answer {\it ``Do not understand''} in task (1) is shown as {\it ``no answer.''} The average of evaluation values is calculated by assigning {\it ``no answer''} a score of 0.
The proposed method outperformed the baseline in terms of relevance to the region and relevance to the answer. The reason is that the proposed method specifies a target object to the LSTM to generate the question, whereas the baselines do not consider any target.
Regarding the number $k$ of class labels used to select the target word, the average score is higher when $k=3$, but the ratio of the highest score 5 is higher when $k=2$.
Also, the proportion of {\it ``no answer''} is higher when $k=2$. This means that, when $k=3$, the target word becomes more generic and the relevance with the target region is less clear than when using $k=2$. On the other hand, a value of $k=2$ is more likely to specify a wrong target word for the visual question generation.

Table~\ref{table:unknown} shows the number of generated questions that successfully allowed acquiring information on unknown objects. We consider successful questions whose answers were not included in the known class of the unknown classifier neither in their hypernym, and whose relevance score in task (2) was 4 or more.
We obtained the highest number of successful questions using our method with $k=2$, since the selected target word is more concrete.
On the other hand, our method with $k=3$ generates questions for a more generic target word, and thus, it does not necessarily get the expected answer, but is still partly related to the target region.

We can conclude that the proposed method can successfully generate questions that allow acquiring information about unknown objects in an image.

\section{Conclusions}\label{conclusion}
In this paper, we presented a novel visual question generation (VQG) task to acquire class information of unknown (i.e., not learned previously) objects in the image, and proposed a method that automatically generates questions that target unknown objects in the image.
To the best of our knowledge, this is the first research that approaches acquiring unknown information via VQG.
The evaluation of our method shows that it can successfully acquire class information of unknown objects from humans. We believe this research will help other researchers in tackling this novel task.

Our future work includes feeding back the acquired information about the unknown object to the system, and learning it as new knowledge.
For example, our method could be combined with recent works in few-shot learning and incremental learning to re-train the classifier with the new class.
In addition, it is considered that the answers from humans will be noisy, so a system that can use noisy answers for re-training is necessary.
If the answer obtained from humans is not the expected, it can be useful to generate multiple questions as necessary.

\section{Acknowledgement}
This work was supported by JST CREST Grant Number JPMJCR1403, Japan.

\bibliographystyle{splncs}
\bibliography{egbib}
\end{document}